\documentclass[%
reprint,
nofootinbib,
 amsmath,amssymb,
 aps,
superscriptaddress,
showkeys,
longbibliography
]{revtex4-1}

\usepackage{xr}
\usepackage{tabularx}
\usepackage{graphicx}
\usepackage{dcolumn}
\usepackage{bm}
\usepackage{microtype}
\usepackage{color}
\usepackage{multirow}
\usepackage{array, booktabs}
\usepackage{enumitem}
\usepackage{threeparttable}
\usepackage{url}
\usepackage{breakurl}
\usepackage[breaklinks, hidelinks, colorlinks=true]{hyperref}

\usepackage{algorithm}
\usepackage{algpseudocode}

\makeatletter
\newcommand*{\addFileDependency}[1]{
  \typeout{(#1)}
  \@addtofilelist{#1}
  \IfFileExists{#1}{}{\typeout{No file #1.}}
}
\makeatother

\begin{document}


\title{A new dataset for measuring the performance of blood vessel segmentation methods under distribution shifts}


\author{Matheus Viana da Silva}
\affiliation{Department of Computer Science, Federal University of S\~ao Carlos, S\~ao Carlos, SP, Brazil}

\author{Natália de Carvalho Santos}
\affiliation{São Carlos Institute of Physics, University of São Paulo,  
São Carlos, SP, Brazil}

\author{Julie Ouellette}
\author{Baptiste Lacoste}
\affiliation{Department of Cellular and Molecular Medicine, Faculty of Medicine, University of Ottawa, Ottawa, ON, Canada}
\affiliation{Neuroscience Program, Ottawa Hospital Research Institute, Ottawa, ON, Canada}

\author{Cesar H. Comin}
\email[Corresponding author: ]{comin@ufscar.br}
\affiliation{Department of Computer Science, Federal University of S\~ao Carlos, S\~ao Carlos, SP, Brazil}

\date{\today}

\begin{abstract}
Creating a dataset for training supervised machine learning algorithms can be a demanding task. This is especially true for medical image segmentation since one or more specialists are usually required for image annotation, and creating ground truth labels for just a single image can take up to several hours. In addition, it is paramount that the annotated samples represent well the different conditions that might affect the imaged tissues as well as possible changes in the image acquisition process. This can only be achieved by considering samples that are typical in the dataset as well as atypical, or even outlier, samples. We introduce VessMAP, a heterogeneous blood vessel segmentation dataset acquired by carefully sampling relevant images from a larger non-annotated dataset. A methodology was developed to select both prototypical and atypical samples from the base dataset, thus defining an assorted set of images that can be used for measuring the performance of segmentation algorithms on samples that are highly distinct from each other. To demonstrate the potential of the new dataset, we show that the validation performance of a neural network changes significantly depending on the splits used for training the network.
\end{abstract}

\keywords{dataset sampling, efficient annotation, diverse dataset, segmentation, blood vessel}

\maketitle

\section{Introduction}


The performance of neural networks has dominantly been measured using metrics such as classification or segmentation accuracy, precision, recall, and the area under the ROC curve. However, recent studies have shown the dangers of only considering such globally-averaged metrics~\cite{Shit2020, Mosinska2018, DaSilva2022} that provide only an aggregated, summarized, view of the performance of machine learning algorithms on datasets with sometimes millions of images. Such an approach may hide important biases of the model \cite{DaSilva2022}. For instance, for medical images, a 95\% accuracy is usually considered a good performance. But what about the remaining 5\%? It is usually unrealistic to expect models to reach 100\% accuracy, but the samples that are not correctly processed by a neural network may hide important biases of the model. These concerns led to the definition of new approaches and metrics that can aid the interpretation of black box models \cite{Chaddad2023}. 

For the task of segmentation in medical images the detection of relevant structures is usually only the first step of a more elaborate procedure for measuring relevant properties such as size \cite{DePMendes2021}, regularity \cite{Perez-Beteta2018}, length \cite{Paetzold2021, FreitasAndrade2022}, and curvature \cite{Krestanova2020, FreitasAndrade2022} of the imaged structures. Therefore, systematic segmentation mistakes might lead to undetected errors when characterizing samples for clinical diagnoses \cite{Reinke2021} and research purposes \cite{Shit2020}. An important cause of such systematic errors can be the presence of samples having characteristics that occur with low frequency in a dataset. This can happen due to additional, unexpected, noise during image acquisition, variations in tissue staining, image artifacts, or even the presence of structures that are anatomically different than what was expected. Assuming for illustration purposes that the data is normally distributed, a machine learning model having good performance around the peak of the distribution will tend to have good average accuracy measured for the whole dataset, even if it cannot correctly classify or segment images that are around the tail of the distribution \cite{Gupta2019}, which might be important for downstream analyses. Notice that this discussion does not necessarily only involves outlier images, but any image occurring with low probability.

We argue that a machine learning model should have good performance, or even be directly optimized, on both prototypical and atypical samples. This focus can lead to models that are more robust to samples located in a sparsely populated region of the feature space of the dataset. In addition, it might also lead to models that generalize better to out-of-distribution samples as well as to new datasets. With these aspects in mind we create a new dataset which we call \emph{Feature-Mapped Cortex Vasculature Dataset} (VesMAP). The dataset is designed to be as heterogeneous as possible by including samples having very different characteristics from each other. To this end, we develop a simple and intuitive sampling methodology to select a subset of 100 images from a base non-annotated dataset containing 18279 image patches. The selected samples were manually annotated with pixel-wise accuracy and represent the final VessMAP dataset. 

The dataset allows the creation of training and validation splits with images having different characteristics such as contrast and blood vessel density. For instance, the dataset allows training on images having very low contrast and validating on high-contrast images. We show that different splits of VessMAP lead to very different training and validation results. Thus, we expect the dataset to be useful for the development of new segmentation algorithms that are robust under distribution shift of the data. We also discuss applications of the dataset for few-shot and active learning. 


\section{Related works}

Table \ref{tab:datasets} shows a summary of the main blood vessel datasets used in the literature. Most of the datasets have images from the retina. Few datasets are associated with microscopy images. More importantly, to our knowledge, none of the datasets were specifically designed to maximize the diversity of the samples. The diversity on some datasets tend to come as a proxy from the inclusion of healthy and abnormal tissue. For instance, samples in the DRIVE dataset contain diabetic retinopathy, which generates blood vessels having abnormal characteristics and image artifacts such as exudates that are not related to blood vessels. Still, most blood vessels tend to have a well-defined geometry and texture in all samples of the dataset. Thus, it becomes a simple task for a segmentation algorithm to generalize to new unseen samples from the same dataset. It is not surprising that many methods can reach an accuracy larger than 0.94 on the DRIVE dataset \cite{Kovacs2022}.

\newcolumntype{P}[1]{>{\centering\arraybackslash}p{#1}}
\renewcommand{\arraystretch}{1.3} 

\renewcommand{\arraystretch}{1.3} 
\begin{table*}[!ht]
 \centering
    \begin{threeparttable}[b]
        \begin{tabularx}{0.8\textwidth} { 
            >{\raggedright\arraybackslash} m{0.2\textwidth} 
            >{\centering\arraybackslash}X 
            >{\raggedleft\arraybackslash}X }
            \toprule
            Dataset & Anatomical Region & Imaging Technique \\ \hline
            DRIVE~\cite{StaalDRIVE}\tnote{1} & \multirow{13}{*}{Retina} & \multirow{13}{*}{Color Fundus Photography} \\
            STARE~\cite{HooverSTARE}\tnote{2} & & \\
            CHASEDB1~\cite{FrazCHASEBD1}\tnote{3} & & \\
            HRF~\cite{OdstrcilikHRF} & &\\
            INSPIRE-AVR~\cite{NiemeijerINSPIRE-AVR}\tnote{4} & & \\
            IMAGERET~\cite{KauppiDIARETDB0,KauppiDIARETDB1}\tnote{5} & & \\
            MESSIDOR~\cite{DecenciereMESSIDOR}\tnote{6} & & \\
            VICAVR~\cite{VICAVR} & & \\
            ROC~\cite{NiemeijerROC} & & \\
            DRIONS DB~\cite{CarmonaDRIONSDB}\tnote{7} & & \\
            DR HAGIS~\cite{HolmDRHAGIS}\tnote{8} & & \\
            RET-TORT~\cite{ForacchiaRETTORT}\tnote{9} & &\\ 
            WIDE~\cite{EstradaWIDE} & & \\ \cmidrule{1-3}
            VAMPIRE~\cite{PerezVAMPIRE} & Retina & Ultra-Wide Field-of-View Fluorescein Angiogram \\ \cmidrule{1-3}
            IOSTAR~\cite{ZhangIOSTARRCSLO} & \multirow{2}{*}{Retina} & Scanning Laser Ophthalmoscopy \\
            RC-SLO~\cite{ZhangIOSTARRCSLO} & & \\ \hline
            Vascular Model Repository~\cite{WilsonVASCULAR} & Aorta; Cerebral, coronary aortofemoral and pulmonary arteries & CTA and MRA Scans \tnote{*} \\
            VESSEL12~\cite{RudyantoVESSEL12} & Lung & Computed Tomography \\
            3D-IRCADb-01~\cite{SolerIRCADb}\tnote{10} & Liver & Computed Tomography \\
            ASOCA~\cite{GharleghiASOCA} & Coronary Arteries & Cardiac CTA \\
            Vascular Synthesizer \cite{HamarnehVascuSynth} & Synthetic Vessels & - \\ \hline
            VesSAP~\cite{TodorovVESSAP} & \multirow{3}{*}{Mouse Brain Vasculature} & \multirow{3}{*}{3D Light-Sheet Microscopy} \\
            TubeMap~\cite{KirstTUBEMAP} & & \\
            Di Diovanna et al.~\cite{DiGiovanna2018} & & \\
            \bottomrule
        \end{tabularx}
        
        \begin{tablenotes}
            \item[1] \url{https://drive.grand-challenge.org/}
            \item[2] \url{https://cecas.clemson.edu/~ahoover/stare/}
            \item[3] \url{https://blogs.kingston.ac.uk/retinal/chasedb1/}
            \item[4] \url{https://medicine.uiowa.edu/eye/inspire-datasets}
            \item[5] \href{https://www.kaggle.com/datasets/nguyenhung1903/diaretdb1-standard-diabetic-retinopathy-database}{DIARETDB0} and \href{https://www.kaggle.com/datasets/nguyenhung1903/diaretdb1-v21}{DIARETDB1}
            \item[6] \url{https://www.adcis.net/en/third-party/messidor/}
            \item[7] \url{http://www.ia.uned.es/~ejcarmona/DRIONS-DB.html}
            \item[8] \url{https://personalpages.manchester.ac.uk/staff/niall.p.mcloughlin/}
            \item[9] \url{https://bioimlab.dei.unipd.it/Retinal\%20Vessel\%20Tortuosity.htm}
            \item[10] \url{https://www.ircad.fr/research/data-sets/liver-segmentation-3d-ircadb-01/}
            \item[*] CTA: Computed Tomography Angiography. MRA: Magnetic Resonance Angiography.
        \end{tablenotes}
        \caption{Summary of the main blood vessel datasets used in the literature. If a dataset cannot be found using the referenced paper, we added its link as a footnote.}
    \end{threeparttable}
    \label{tab:datasets} 
\end{table*}

Regarding microscopy images, all datasets found by our survey include very few samples. Usually, very large 3D volumes are annotated in a semi-supervised fashion. They contain large amounts of vessels, but represent a single individual and image acquisition procedure. Therefore, most vessels have similar appearance and it becomes difficult to measure the generalization capability of segmentation methods. With these limitations in mind, we created a dataset that was specifically designed to include blood vessels having very different characteristics. 

The creation of the dataset involved the development of a methodology for selecting relevant samples for annotation. A concept that is similar to the developed methodology is the so-called \emph{coreset} \cite{Guo2022}. The aim of a coreset is to select a subset of samples that can optimally represent the whole dataset. Many different methodologies and criteria were developed for defining relevant coresets \cite{Zheng2019, Adhikari2021, Guo2022}. Indeed, the subset defined by our methodology can be associated with a coreset, but in our case, the aim of the methodology and the approach used differs markedly from the usual definition of a coreset. The aim of our methodology is not focused on accurately representing the underlying distribution of the data, or on preserving the accuracy of a machine learning algorithm, but on providing a relevant dataset for training machine learning algorithms while avoiding the underrepresentation of atypical samples. In addition, many coreset methodologies use a surrogate neural network to estimate latent features or to estimate a degree of uncertainty about each sample, while our methodology is more general in the sense that any set of features obtained from the samples can be used. Furthermore, many related studies consider a similarity metric for selecting relevant samples \cite{Zheng2019, Adhikari2021}, which is a degenerate metric and therefore cannot provide a full representation of the data distribution.







\section{Materials and methods}
\label{sec:methodology}

\subsection{Blood vessel microscopy base images}

We start from a collection of 2637 confocal microscopy images of mouse brain vasculature. The images were acquired under different experimental conditions in different works published in the literature \cite{Lacoste2014, Gouveia2017, Ouellette2020}. Conditions include control animals, animals that have suffered a deletion of chromosome 16p11.12, animals that have experienced sense deprivation or sense hyperarousal, samples from stroke regions, and also from different stages of mouse development. The images have sizes from $1376\times 1104$ to $2499\times 2005$ pixels, totaling around 3.8GB of data. 

The dataset is interesting because it has a considerable variety of characteristics of blood vessels. In addition, the images represent samples obtained from hundreds of different animals and experimental conditions. This makes it an excellent dataset for training machine learning algorithms for blood vessel segmentation. But training supervised algorithms requires the manual annotation of the blood vessels.

After annotating a few samples, we estimated that each image in the dataset takes roughly 12 hours to fully annotate. Therefore, it is unfeasible to annotate the whole dataset, and a subset of samples needed to be selected.  Our objective was to select a diverse set of samples containing both prototypical and atypical samples, so that it would be possible to create useful train/val splits for quantifying the performance of segmentation algorithms under challenging distribution shifts between the splits. To this end, a sampling methodology was developed to select appropriate samples. 

\subsection{Sampling methodology}

\begin{figure}[]
    \centering
    \includegraphics[width=0.8\linewidth]{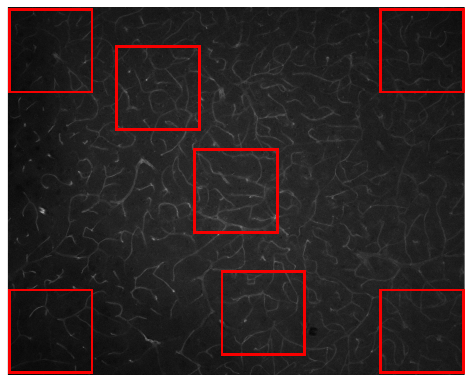}
    \caption{An example of seven regions extracted from a single sample. The four corners along with the central region can capture most of the illumination inhomogeneities that may occur due to uneven illumination of the samples. Besides these five regions, two additional random regions are also drawn for each image.}
    \label{fig:janelas}
\end{figure}

Each image in the base dataset may include illumination inhomogeneities, changes in contrast, different levels of noise, as well as blood vessels having distinct characteristics (e.g., caliber, tortuosity, etc). Thus, from the original dataset, we generated a new set of images, each having a size of 256$\times$256 pixels. These smaller images were generated by extracting 256$\times$256 patches from the original images. As shown in Figure \ref{fig:janelas}, seven regions were extracted from each image. The seven regions were extracted in key areas of each image, with four windows in each of the corners of the image, a central window, and two windows at random positions. The latter two may overlap with the other windows. Windows that did not contain a satisfactory number of blood vessel segments were removed. The total size of the resulting dataset is 18279 images. This new dataset was used in the remainder of the sampling procedure. 

The methodology developed to sample relevant images has three steps: (a) dataset mapping to an n-dimensional feature space, (b) generation of a discrete representation of the feature space, and (c) selection of points from the feature space representation. We explain each of these steps in the following sections.


\subsubsection{Dataset mapping}
\label{sec:dataset_mapping}

\begin{figure*}[]
    \centering
    \includegraphics[width=0.8\linewidth]{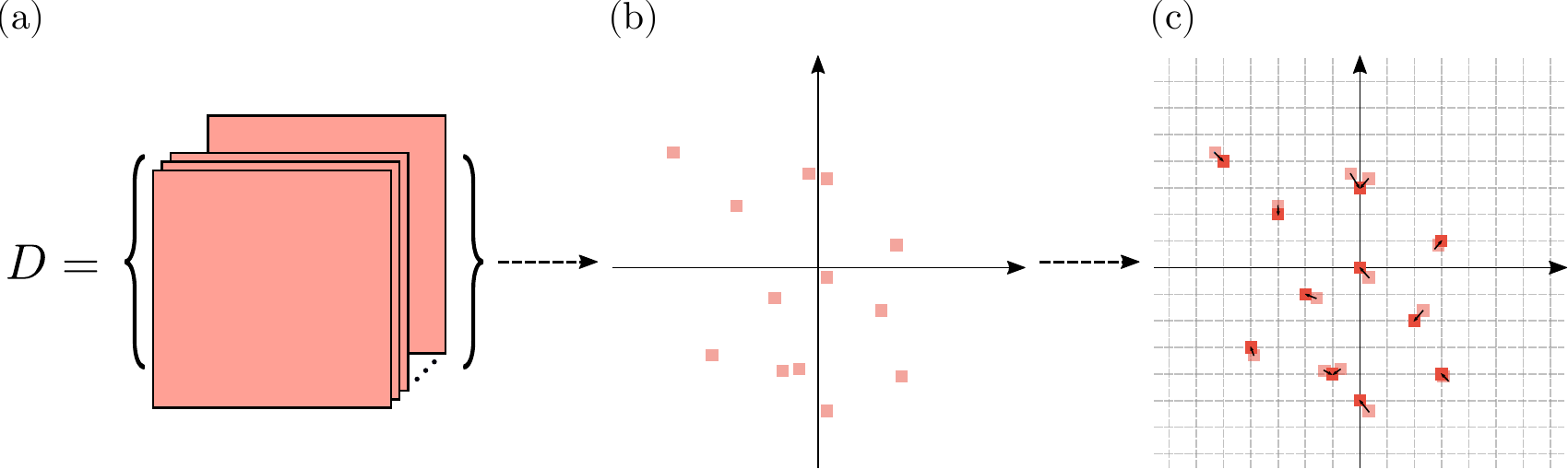}
     \caption{Representation of the mapping procedure applied to a set $D$ of samples, followed by the feature space discretization. Here, we consider $D$ as a set of blood vessel images. In this example, each image of $D$ is mapped to a 2-d position in the new feature space (b). In (c), the mapped points (light-red points) are moved to a new position (red points) within a regular grid defined by Equation \ref{eq:resample}.}
    \label{fig:data_mapping}
\end{figure*}

We represent the dataset as $D = \{\delta_1, \delta_2, ..., \delta_n\}$ where $n$ is the number of samples. Given a function $f: \delta_i \to \Vec{p_i}$ that maps a sample $\delta_i$ to a vector $\Vec{p_i}$ with dimension $d$, the dataset is mapped to a feature space as a $n \times d$ matrix, which we call $D_{mapped}$. Each line of this matrix therefore represents the features of a sample $f(\delta_i)$. Figure \ref{fig:data_mapping} illustrates this procedure. Each sample in $D$ (Figure \ref{fig:data_mapping}(a)) is mapped to a point in the new feature space (Figure \ref{fig:data_mapping}(b)). In Section \ref{sec:case_example} we describe the features used to create the mapping. 


\subsubsection{Feature space discretization}
\label{sec:feature_space_discretization}

A regular grid was defined in the feature space and each data point was mapped to the nearest point in this grid. It is useful to first normalize the values of $D_{mapped}$ to remove differences in the scale of the features. The z-score of the features was used. After normalizing $D_{mapped}$, the values were mapped to the discrete grid. This was done by defining a scale $\nu$ that set the size of each grid cell and calculating

\begin{equation}
    \label{eq:resample}
    \displaystyle D_{grid} = floor\left(\frac{D_{mapped}}{\nu}\right),
\end{equation}
where $floor$ is the floor function. As shown in Figure \ref{fig:data_mapping}(c), this operation ensures that each value of $D_{grid}$ lies within a regular grid. Note that, as a consequence of undersampling, we expect multiple data points to fall in the same grid position, this is one of the key properties of the method that will allow a uniform sampling of the data.

After feature space discretization, we generate a sparse set of points representing an estimation of the possible values that can be obtained in the feature space. We call this set the \emph{sampling set} of the feature space. This procedure works as follows. An n-dimensional discrete hypersphere $S$ with radius $r$ (in grid units) centered on each data point is defined. This hypersphere is translated to each data point position. The union of the calculated hypersphere positions of all points defines the sampling set $D_{sset}$. The general appearance of $D_{sset}$ is depicted by the blue points of Figure \ref{fig:drawing}.



\subsubsection{Uniform selection of points}
\label{sec:sampling}

\begin{figure}
    \centering
    \includegraphics[width=0.6\linewidth]{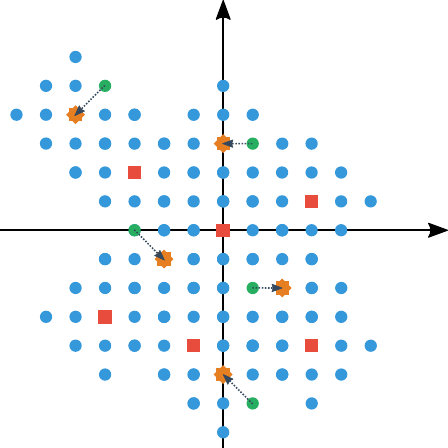}
    \caption{Illustration of the proposed sampling protocol. $k$ random points (green dots) are drawn from the sampling set $D_{sset}$ (blue dots). The subset of sampled data points is defined by the data points that are closest to each drawn point (orange stars). Red squares represent the remaining data points that were not selected.}
    \label{fig:drawing}
\end{figure}

The samples are selected by first drawing a set of points from the sampling set $D_{sset}$. As illustrated in Figure \ref{fig:drawing}, we draw from $D_{sset}$ $k$ points with uniform probability (green dots in Figure~\ref{fig:drawing}). For each point drawn, the closest data sample is identified using the Euclidean distance. If the same data sample is obtained more than once, a new point is drawn from $D_{sset}$ until $k$ unique data samples are obtained. The final set of data samples (orange stars in Figure~\ref{fig:drawing}) is represented as $D_{sampled}$.

A uniform sampling of $D_{sset}$ allows the selection of prototypical and atypical samples from the dataset with equal probability. Nevertheless, a single realization of the sampling may lead to distortions such as the selection of many samples at similar regions of the space or the creation of large regions with no samples selected. This is due to random fluctuations in the sampling process. To amend this, we define a metric called Farthest Unselected Point (FUS), that punishes sampled subsets with large gaps between the selected points.

Let $D_{sampled}$ be the set of sampled data points from $D_{grid}$, and $\neg D_{sampled}$ the set of points from $D_{grid}$ that were not selected in the sampled subset. For each data point in $\neg D_{sampled}$, the Euclidean distance to the closest point in $D_{sampled}$ is obtained. The FUS metric is defined as the largest calculated distance among all points in $\neg D_{sampled}$.
Sampled subsets leading to low values of the FUS metric should be preferred, since it avoids the creation of large regions of the feature space with no samples. In our experiments, we found that  minimizing FUS for 1000 different subsets covered a good amount of subset possibilities.



\subsection{Creating a feature-mapped blood vessel dataset}
\label{sec:case_example}

Given that the images from our base dataset were used in previous works, each sample has a respective segmentation that was obtained using a semi-supervised methodology. This methodology is based on the adaptive thresholding of the original images, where the threshold was selected manually for each image. The full details on the segmentation procedure are described in \cite{FreitasAndrade2022}. Using the semi-supervised segmentation, the following features were used to characterize the samples: blood vessel contrast, level of Gaussian noise, blood vessel density, and medial line heterogeneity. 

The blood vessel contrast is related to the average difference in intensity between the vessels and the background of the image. The greater the contrast, the easier it is to detect the vessels. It can be measured using the original image of the vessels and the respective thresholded image containing an estimation of the pixels belonging to the vessels. The contrast is calculated as 

\begin{equation}
    C = \frac{\bar{I}_v}{\bar{I}_f},
    \label{contrast}
\end{equation}
where $\bar{I}_v$ and $\bar{I}_f$ are the mean intensities of, respectively, the pixels belonging to the blood vessels and the background of the image.

The signal-to-noise level of the images can be estimated in different ways. We investigated different definitions and used the method that was the most compatible with a visual inspection of the images. The method proposed in \cite{Donoho1994} was used. It assumes a noise with normal distribution and uses wavelets to identify the most likely standard deviation of the noise component. To prevent the method from capturing vessel variation, only the background of the image was used for the estimation. 

Blood vessel density is defined as the total length of blood vessels in an image divided by the image area. To do this, we first apply a skeletonization algorithm to extract the medial lines of the vessels \cite{Palagyi1998}. The total length of vessels is calculated as the sum of the arc-lengths of all vessel segments.

The last metric, which we call medial line heterogeneity, measures the illumination changes in the vessel lumen. To calculate this metric, we first blur the image using a gaussian filter with unit standard deviation to remove extreme values. The medial line heterogeneity is calculated as the standard deviation of the pixel values along the medial lines of this blurred image. The medial lines considered are the same ones used for the blood vessel density metric.

We observed that the medial line heterogeneity tends to be correlated with the average intensity of the blood vessels. In order to remove this dependency, the medial line heterogeneity as well as the average intensity of the medial lines were calculated for all images in the dataset. Then, a straight line fit $h_m = a*m + b$ was applied to the calculated values, where $m$ is the average intensity and $h_m$ is the expected medial line heterogeneity associated with $m$. Next, a normalized medial line heterogeneity was defined as $\tilde{h} = h - h_m$, where $h$ is the medial line heterogeneity calculated for an image.

The four metrics were used for mapping the dataset to a 4-d feature space. As mentioned before, the dataset contains 18279 images. Hence, the whole dataset was mapped to a 18279 $\times$ 4 matrix. In the feature space discretization step (Section \ref{sec:feature_space_discretization}), we used a scale of $\nu=10$. The hypersphere was generated with a radius equal to four times the grid space resolution. We decided to select $k = 100$ images for annotation. To avoid data leakage, an additional restriction that prevented the selection of samples from the same image was used. 

\section{Results}

\subsection{Dataset heterogeneity}

The sampling approach used to generate the VessMAP dataset should lead to a heterogeneous set of samples. It is difficult to properly measure the heterogeneity of the dataset because it would involve the estimation of the probability density function of the original data, which is not a trivial task and can be influenced by the choice of parameter values. But it is clear that the method should naturally lead to a uniform selection of the samples. This is so because the set $D_{grid}$ (defined in Section \ref{sec:feature_space_discretization}) represents an estimation of the domain of the probability density function of the data, and this domain is being sampled uniformly.

\begin{figure}[]
    \centering
    \includegraphics[width=0.8\linewidth]{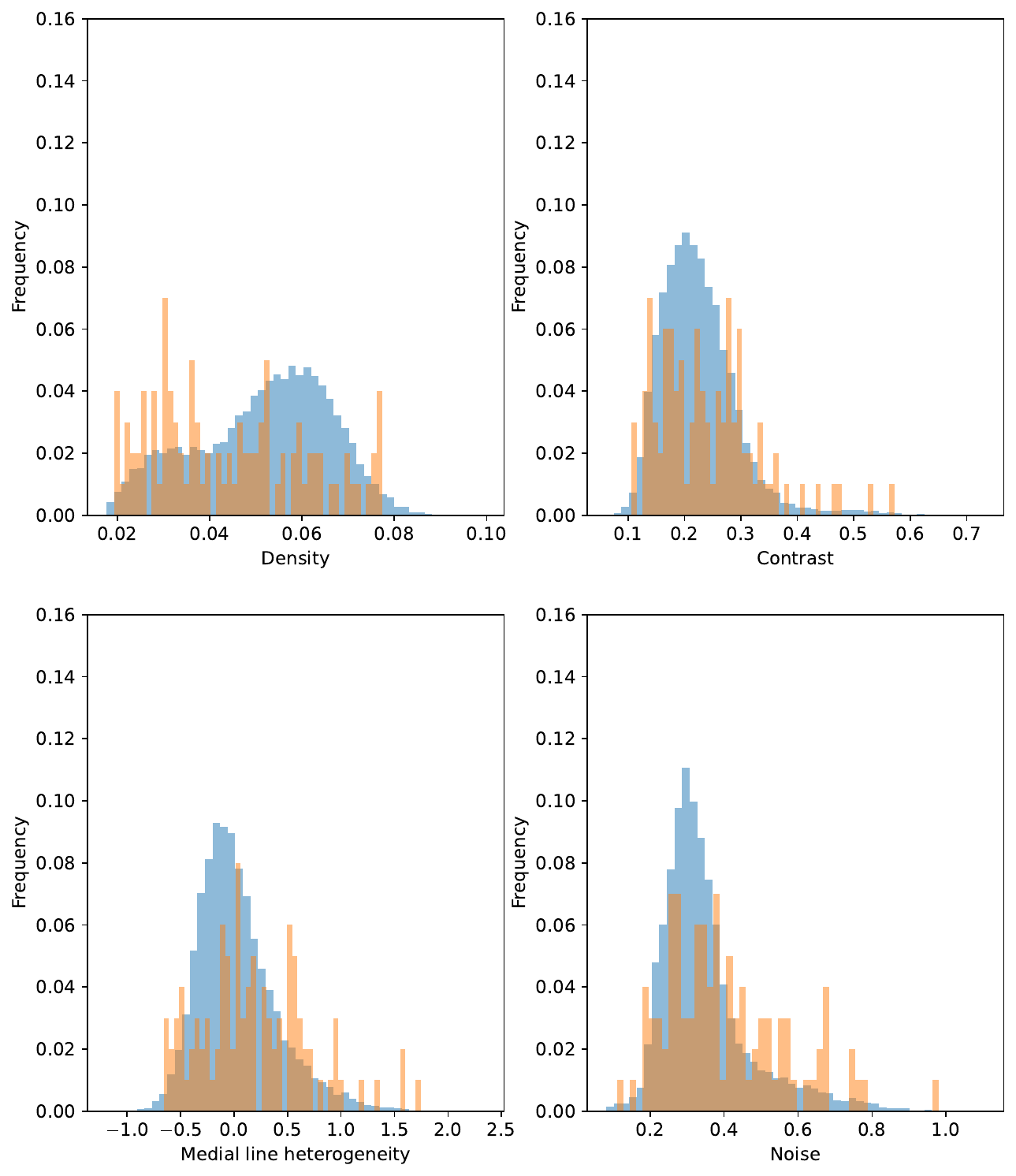}
    \caption{Histograms of the four features calculated from the dataset. Blue bars correspond to the distribution of the full data. Orange bars correspond to the distribution of the sampled subset. Note that the frequencies were normalized by their sum.}
    \label{fig:metrics_histogram}
\end{figure}

One approach to illustrate the characteristics of the sampled images is displayed in Figure \ref{fig:metrics_histogram}, which shows histograms of the four considered features for both the full dataset and the sampled subset. The histograms of individual features are not expected to be uniform since they represent a projection of the original data into one dimension. Still, it can be seen that the histograms of the sampled set tend to represent a slightly flattened version of the histograms of the original data, indicating that a larger priority is being given to atypical samples when compared to the original distribution.

\begin{figure}[]
    \centering
    \includegraphics[width=\linewidth]{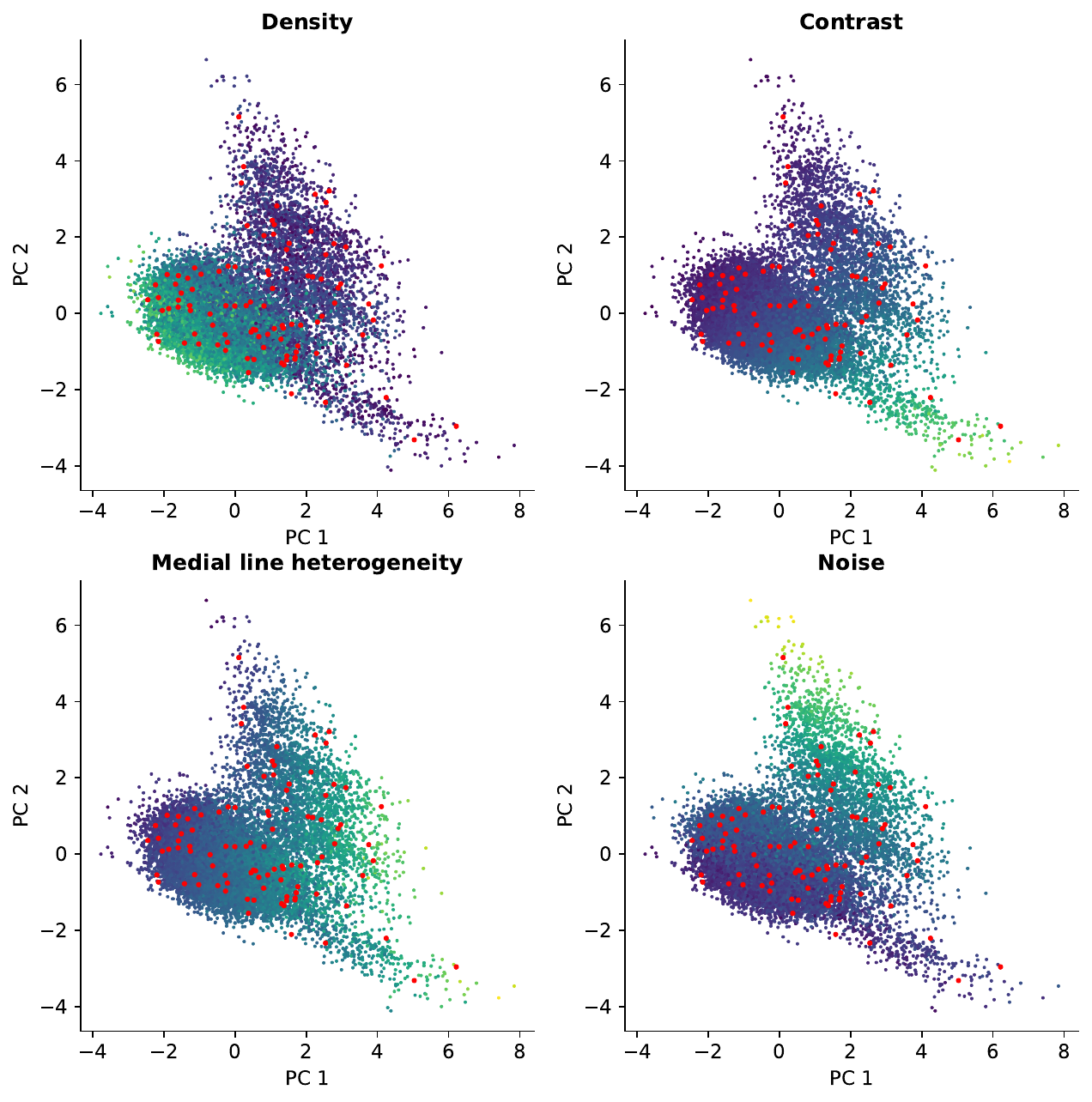}
    \caption{PCA of the blood vessel dataset. The original features with z-score normalization was used. Red points correspond to the sampled subset obtained by the sampling methodology. Other points correspond to the unselected points, with their colors representing the value of each one of the four original metrics: vessel density, contrast, medial line heterogeneity, and image noise.}
    \label{fig:pca}
\end{figure}

A more robust way of visually checking the sampled subset is to visualize the data using Principal Component Analysis (PCA). Using PCA, the original 4-d data can be projected into 2-d with optimal preservation of the variance. Figure \ref{fig:pca} shows the PCA projection of the data. The four plots included in the figure represent the same projection, but the points are colored according to the different features used to characterize the images. The selected samples are shown in red.
It can be noticed that the sampling methodology selects a subset of images that uniformly covers the distribution of the data. Furthermore, as also suggested by the histograms in Figure \ref{fig:metrics_histogram}, the sampling was capable of covering the full range of values of every considered feature.

\begin{figure*}[]
    \centering
    \includegraphics[width=\linewidth]{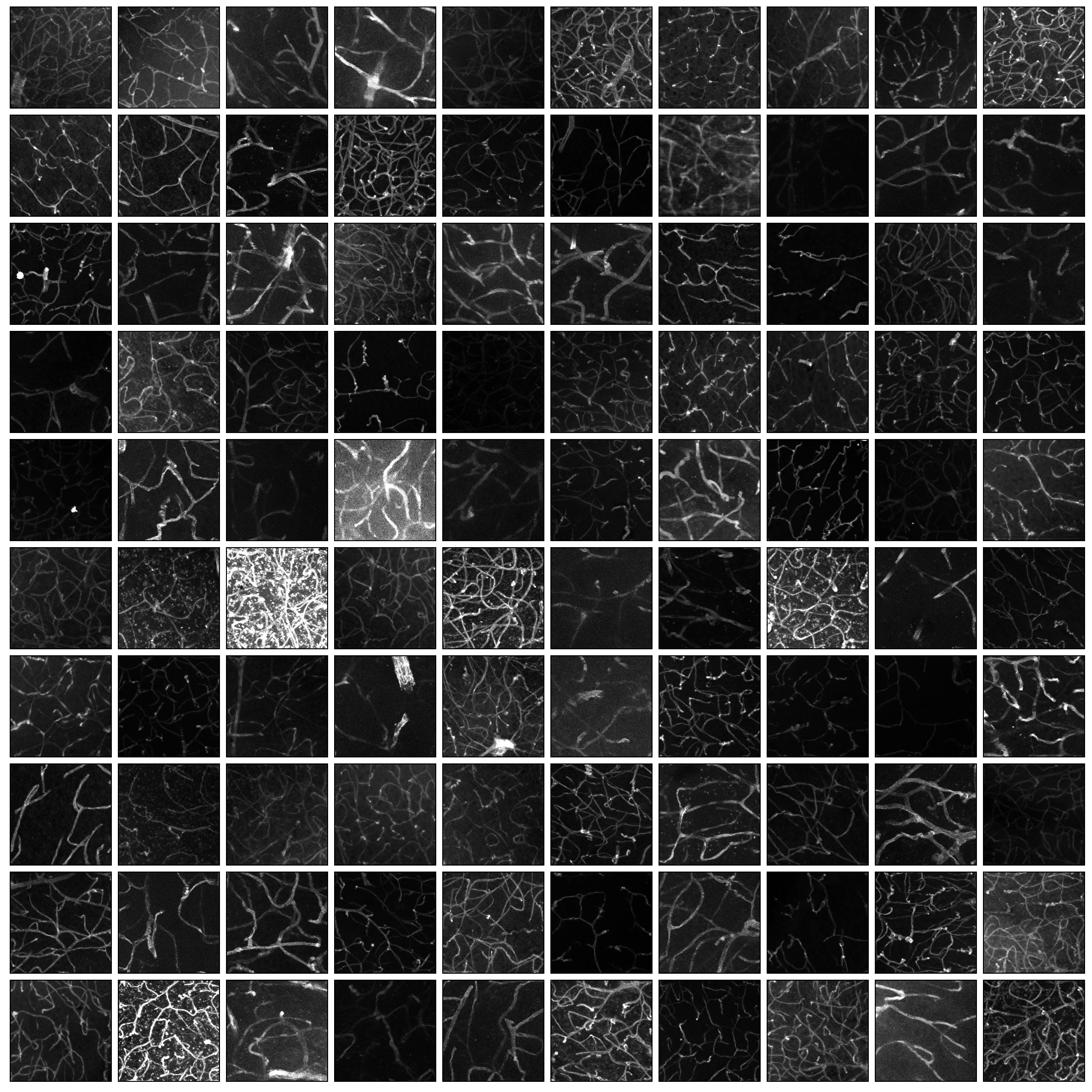}
    \caption{The VessMAP dataset. The images cover a wide range of values in the feature space defined by our four features. Contrast variation and vessel density are the easier features to visually verify. Medial line heterogeneity can be verified by how much the brightness changes longitudinally within a blood vessel. Gaussian noise level is harder to verify visually, but pronounced noise can be observed on some of the brighter images.}
    \label{fig:sampled_images}
\end{figure*}

The subset of images selected by the method is shown in Figure \ref{fig:sampled_images}. The subset indeed contains a heterogeneous set of images covering many different values of the considered features (e.g., low contrast, high vessel density, etc). For instance, some of the samples in the dataset come from animals who suffered hemorrhagic strokes. These samples are very different from the typical samples contained in the base dataset, and they would be largely underrepresented if a sampling following the data distribution was performed. 

We manually labeled each of the 100 images and made the dataset publicly available~\footnote{VessMAP (Feature-Mapped Cortex Vasculature Dataset): \url{https://doi.org/10.5281/zenodo.10045265}}. To account for inter-annotator variability, 20 samples were labeled by two annotators. The repository includes manually annotated binary labels, their skeletons (calculated by the Palágyi-Kuba algorithm \cite{Palagyi1998}), and the metrics for each sample (as described in section \ref{sec:case_example}) -- which were calculated using the manual annotations. We verified that the metrics calculated from the manual annotations have a strong correlation with the metrics calculated using the labels obtained from the semi-supervised segmentation algorithm. This evidences the quality of the algorithm in providing useful metrics to map the dataset into a feature space. We expect the VessMAP Dataset to be useful for future studies regarding the influence of image and tissue characteristics on the generalization capability of segmentation algorithms.

\subsection{Neural network performance on VessMAP splits}

To evaluate the potential of VessMAP to generate data splits that are challenging for neural networks, we generated four different splits based on the features used for creating the dataset: blood vessel density, contrast, medial line heterogeneity, and noise estimation. For each feature, we selected 20 of the samples with the lowest and highest values and trained a segmentation Convolutional Neural Network (CNN) using two configurations: (i) training with samples that have the lowest feature values --lowest split-- and evaluating with samples that have the highest feature values --highest split--, and (ii) training with the highest split and evaluating with the lowest split. We chose to use 20 images because this is a common split size for well-known blood vessel datasets, such as DRIVE \cite{StaalDRIVE} and STARE \cite{HooverSTARE}. The idea behind this experiment is to test whether we can use VessMAP to generate splits that challenge the generalization capability of CNNs.

The chosen CNN architecture for this experiment was a U-Net with residual layers. This architecture is often employed for the segmentation of blood vessels \cite{Lian2021, Sule2022, Pal2024} and other biological structures \cite{Alom2019}, hence, it is a representative architecture in current research of automatic segmentation and morphometry of microscopic images. For each training/evaluation split, we trained the network for 1000 epochs. The training was carried out using the Cross Entropy as the loss function, and the Polynomial Learning Rate (power = 0.9) as the learning rate scheduler -- which decays the initial learning rate (0.01) almost linearly.

\begin{figure}[!h]
    \centering
    \includegraphics[width=\linewidth]{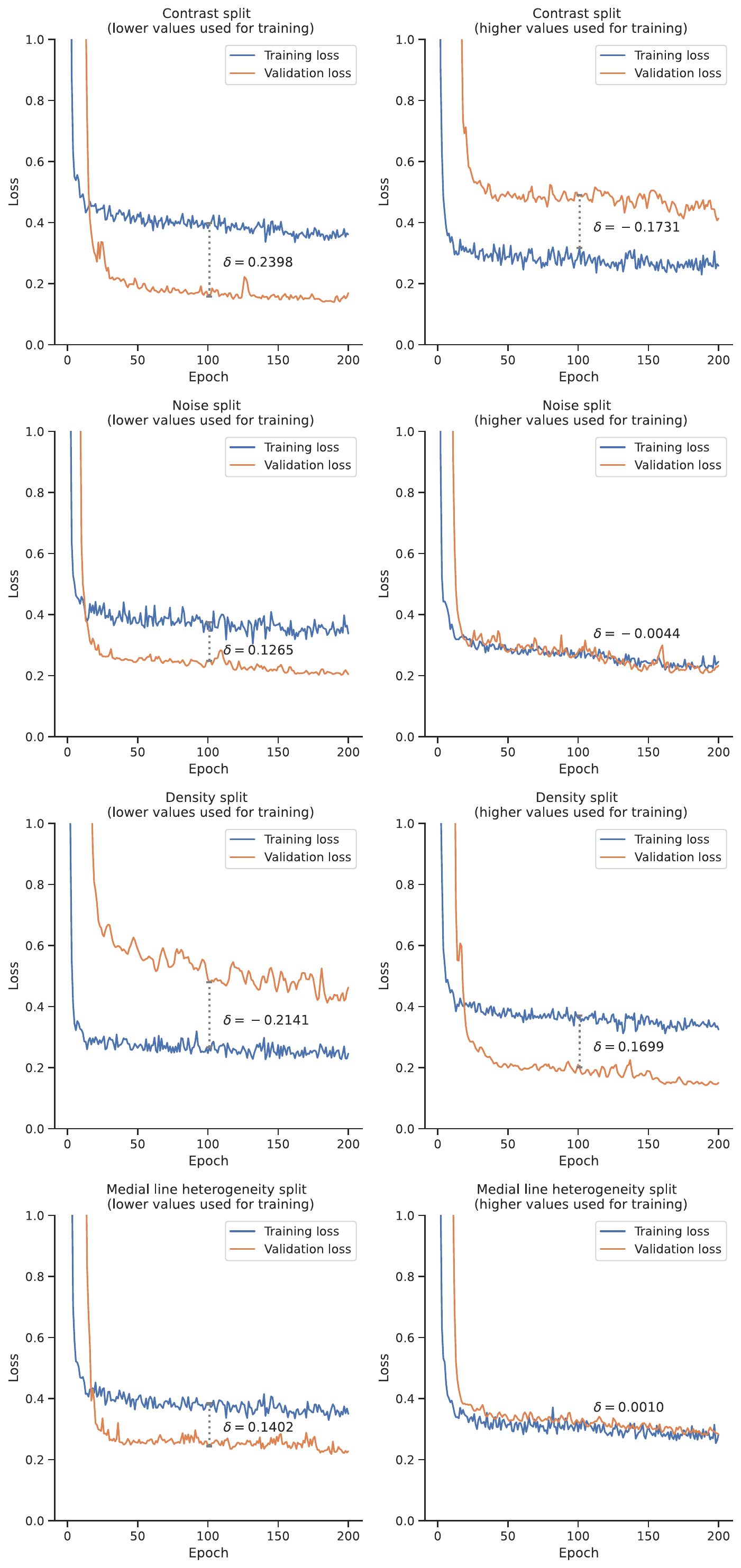}
    \caption{Train and validation loss curves for eight different splits that were generated using the VessMAP metadata. Negative $\delta$ values indicate that the CNN does not generalize well to the validation data. Positive $\delta$ values indicate that the validation samples are easier to segment than the training samples.}
    \label{fig:splits}
\end{figure}

Figure \ref{fig:splits} presents the loss curves of the eight training setups (two split configurations for each metric). First, notice that we only plotted the loss curves of a portion of the 1000 epochs. In our analysis, we evaluate the distance between the training and validation loss curves as a metric of how well the CNN generalized for out-of-distribution data, and since the loss curves remained equally distant for the remaining 800 epochs (for all splits), we consider the first 200 epochs to be sufficiently informative. We define $\delta$ as the difference between the validation loss and the training loss at a specific epoch. Here, we calculate $\delta$ at epoch 100.

For the splits using the contrast feature, when training with samples having low contrast, the network generalizes well for new data having high contrast. This behavior can be attributed to the fact that high-contrast images are less challenging and, if the network learns how to properly segment low-contrast images, it tends to handle well high-contrast images. The opposite behavior occurs when we invert the training and validation sets. When training with high-contrast images, the CNN could not generalize towards low-contrast data, yielding a negative $\delta$. The same behavior can be observed for the blood vessel density splits. Notice that a negative $\delta$ indicates that using specifically low-density samples as the training set yields low generalization towards more dense images. The exact opposite happens when training with denser samples. It can also be noticed that the noise and medial line heterogeneity splits that use the highest values for training resulted in similar training and validation loss curves. This indicates that, although the splits are unbalanced regarding feature values, the samples are still diverse enough to allow good generalization.

With this analysis, we suggest two applications of VessMAP metadata. First, one can use the mapped features to generate splits that challenge the generalization capacity of a neural network, yielding negative $\delta$ during training. This kind of split can be used to test or develop new approaches to handle datasets having very distinct samples. In a similar fashion, splits that have positive $\delta$ can be used for developing new active learning methods, where it is useful to identify challenging samples for training networks so as to obtain low validation loss. In both situations, an ideal model should converge the training and validation loss curves, resulting in $\delta \approx 0$.

\section{Conclusion}

Selecting appropriate images from a larger dataset for training machine learning algorithms is an important task. This is because the usual approach is to use as many images as possible. While this approach is relevant for general classification problems, for medical image segmentation, where image annotation can be very costly, the images used must be carefully selected in order to ensure good coverage of different tissue appearances and imaging variations. In addition, it is important that the selected images do not lead to biases in downstream tasks related to tissue characterization. For instance, training segmentation algorithms mostly on prototypical images can lead to incorrect measurements on samples having unusual properties (e.g., very bright or very noisy). 

We developed an intuitive sampling methodology that evenly selects, as best as possible, both typical and atypical samples for creating a rich dataset that can then be annotated and used for training segmentation algorithms. One important property of the method is that it provides an intuitive uniform grid in the feature space that can be used for further analyses. For example, one can study the accuracy of the segmentation on different regions of the grid to identify regions where samples are not being correctly segmented. A robust algorithm should provide good segmentation no matter if a sample is too noisy, bright or dark, if it has low or high contrast or any other variation on relevant image properties. Likewise, expected tissue changes in the samples should not lead to variations in accuracy. 


The methodology was used for creating VessMAP, a dataset containing a heterogeneous set of samples representing many possible variations of image noise and contrast as well as blood vessel density and intensity variance. The dataset is being made available together with the metadata containing the features used for creating the dataset. We showed that different splits of the dataset can lead to largely distinct validation performances. We expect that the dataset will be useful for studies regarding data distribution shift as well as few-shot and active learning methods.

\section*{Funding}
Cesar H. Comin thanks FAPESP (grant no. 21/12354-8) for financial support. M. V. da Silva thanks FAPESP (grant no. 23/03975-4), Google's Latin America Research Awards (LARA 2021), and the Google PhD Fellowship Program for financial support. The authors acknowledge the support of the Government of Canada's New Frontiers in Research Fund (NFRF) (NFRFE-2019-00641).

\def\UrlBreaks{\do\/\do-}

\bibliography{references}

\end{document}